\documentclass[11pt,a4paper]{article}
\pdfoutput=1
\usepackage[hyperref]{emnlp2018}
\usepackage[utf8]{inputenc}
\usepackage{times}
\usepackage{latexsym}
\usepackage{lipsum}
\usepackage{url}
\usepackage{multirow}
\usepackage{changepage}
\usepackage{amssymb}
\usepackage{bm}
\usepackage{amsmath}
\usepackage{graphicx}
\usepackage{siunitx}
\usepackage{caption}
\usepackage{blindtext}
\usepackage{booktabs}
\usepackage[hang,flushmargin]{footmisc} % remove indentation from footnotes

\aclfinalcopy % Uncomment this line for the final submission
\title{IIIDYT at IEST 2018: Implicit Emotion Classification With Deep
Contextualized Word Representations}

\author{Jorge A. Balazs, Edison Marrese-Taylor \and Yutaka Matsuo \\
  Graduate School of Engineering\\
  The University of Tokyo\\
  \{jorge, emarrese, matsuo\}@weblab.t.u-tokyo.ac.jp \\
}

\date{}

\newcommand{\Acronym}[1]{\mbox{#1}}

\begin{document}
\maketitle
\begin{abstract}

In this paper we describe our system designed for the WASSA 2018 Implicit
Emotion Shared Task (IEST), which obtained 2$^{\text{nd}}$ place out of 30 teams
with a test macro F1 score of $0.710$. The system is composed of a single
pre-trained ELMo layer for encoding words, a Bidirectional Long-Short Memory
Network BiLSTM for enriching word representations with context, a max-pooling
operation for creating sentence representations from them, and a Dense Layer for
projecting the sentence representations into label space. Our official
submission was obtained by ensembling 6 of these models initialized with
different random seeds. The code for replicating this paper is available at
\url{https://github.com/jabalazs/implicit_emotion}.

\end{abstract}

\section{Introduction}

Although the definition of emotion is still debated among the scientific
community, the automatic identification and understanding of human emotions by
machines has long been of interest in computer science. It has usually been
assumed that emotions are triggered by the interpretation of a stimulus event
according to its meaning. 

As language usually reflects the emotional state of an individual, it is natural
to study human emotions by understanding how they are reflected in text. We see
that many words indeed have affect as a core part of their meaning, for example,
\textit{dejected} and \textit{wistful} denote some amount of sadness, and are
thus associated with sadness. On the other hand, some words are associated with
affect even though they do not denote affect. For example, \textit{failure} and
\textit{death} describe concepts that are usually accompanied by sadness and
thus they denote some amount of sadness. In this context, the task of
automatically recognizing emotions from text has recently attracted the
attention of researchers in Natural Language Processing. This task is usually
formalized as the classification of words, phrases, or documents into predefined
discrete emotion categories or dimensions. Some approaches have aimed at also
predicting the degree to which an emotion is expressed in text
\cite{wassa_emoint_2017}.

In light of this, the \Acronym{WASSA} 2018 Implicit Emotion Shared Task
(\Acronym{IEST}) \cite{klinger2018iest} was proposed to help find ways to
automatically learn the link between situations and the emotion they trigger.
The task consisted in predicting the emotion of a word excluded from a tweet.
Removed words, or \textit{trigger-words}, included the terms ``sad'', ``happy'',
``disgusted'', ``surprised'', ``angry'', ``afraid'' and their synonyms, and the
task was to predict the emotion they conveyed, specifically sadness, joy,
disgust, surprise, anger and fear.

From a machine learning perspective, this problem can be seen as sentence
classification, in which the goal is to classify a sentence, or in particular a
tweet, into one of several categories. In the case of \Acronym{IEST}, the
problem is specially challenging since tweets contain informal language, the
heavy usage of emoji, hashtags and username mentions.

In this paper we describe our system designed for \Acronym{IEST}, which obtained
the second place out of 30 teams. Our system did not require manual feature
engineering and only minimal use of external data. Concretely, our approach is
composed of a single pre-trained \Acronym{ELMo} layer for encoding words
\cite{peters2018deep}, a Bidirectional Long-Short Memory Network
(\Acronym{BiLSTM}) \cite{graves2005framewise, graves2013speech}, for enriching
word representations with context, a max-pooling operation for creating sentence
representations from said word vectors, and finally a Dense Layer for projecting
the sentence representations into label space. To the best of our knowledge, our
system, which we plan to release, is the first to utilize ELMo for emotion
recognition.

%  \item Code will be made available

% FIXME: EDIT THIS SECTION TO MAKE PARAGRAPHS MORE COHESIVE
% Sentiment Analysis, also known as Opinion Mining, is a discipline whose
% objective is to automatically identify sentiment in written text
% \cite{balazs2016opinion}.

\begin{figure*}
    \centering
    \includegraphics[scale=0.6]{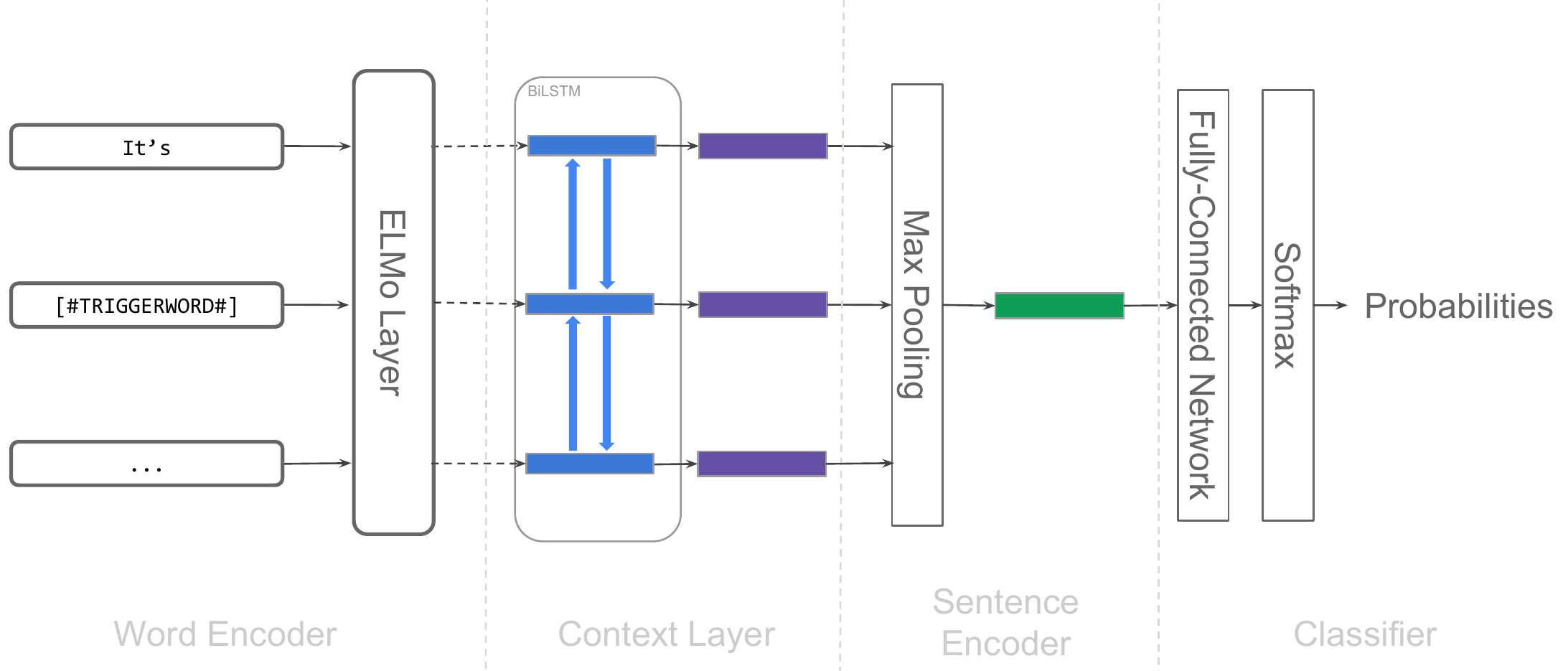}
    \caption{Proposed architecture.}
    \label{fig:architecture}
\end{figure*}

\section{Proposed Approach}

\subsection{Preprocessing}

As our model is purely character-based, we performed little data preprocessing.
Table \ref{table:substitutions} shows the special tokens found in the datasets,
and how we substituted them. 

\begin{table}[!h]
    \centering
    \footnotesize

    \begin{tabular}{ll}
        \toprule
        \textbf{Original} & \textbf{Replacement} \\
        \midrule
        \texttt{\footnotesize[\#TRIGGERWORD\#]} & \texttt{\footnotesize\_\_TRIGGERWORD\_\_} \\
        \texttt{\footnotesize@USERNAME} & \texttt{\footnotesize\_\_USERNAME\_\_} \\
        \texttt{\footnotesize[NEWLINE]} & \texttt{\footnotesize\_\_NEWLINE\_\_} \\
        \texttt{\footnotesize http://url.removed} & \texttt{\footnotesize\_\_URL\_\_} \\
        \bottomrule

    \end{tabular}
    \caption{Preprocessing substitutions.}
    \label{table:substitutions}
\end{table}

    Furthermore, we tokenized the text using a variation of the
    \texttt{twokenize.py}\footnote{\tiny\url{https://github.com/myleott/ark-twokenize-py}}
    script, a Python port of the original \texttt{Twokenize.java}
    \cite{gimpel-EtAl:2011:ACL-HLT2011}. Concretely, we created an emoji-aware
    version of it by incorporating knowledge from an emoji
    database,\footnote{\tiny\url{https://github.com/carpedm20/emoji/blob/e7bff32/emoji/unicode_codes.py}}
    which we slightly modified for avoiding conflict with emoji sharing unicode
    codes with common glyphs used in Twitter,\footnote{For example, the hashtag
    emoji is composed by the unicode code points \texttt{U+23 U+FE0F U+20E3},
which include \texttt{U+23}, the same code point for the \texttt{\#} glyph.} and
for making it compatible with Python 3.

% \begin{itemize}
%     \item briefly describe previous Semeval papers~\cite{baziotis2018ntua,
%         duppada2018seernet, abdou2018affecthor} (Should we do this despite not
%         having based our work in any of them?)
%     \item mention previous Wassa papers
%     \item talk about some sentence classification tasks? Works that fall into
%         the broad category of mapping sentences to labels.
%     \item Talk about elmo~\cite{peters2018deep}
% \end{itemize}

\subsection{Architecture}

Figure \ref{fig:architecture} summarizes our proposed architecture. Our input is
based on Embeddings from Language Models (ELMo) by \citet{peters2018deep}. These
are character-based word representations allowing the model to avoid the
``unknown token'' problem. ELMo uses a set of convolutional neural networks to
extract features from character embeddings, and builds word vectors from them.
These are then fed to a multi-layer Bidirectional Language Model (BiLM) which
returns context-sensitive vectors for each input word. 

We used a single-layer BiLSTM as context fine-tuner \cite{graves2005framewise,
graves2013speech}, on top of the ELMo embeddings, and then aggregated the hidden
states it returned by using max-pooling, which has been shown to perform well on
sentence classification tasks \cite{conneau2017supervised}.

Finally, we used a single-layer fully-connected network for projecting the
pooled BiLSTM output into a vector corresponding to the label logits for each
predicted class.

\subsection{Implementation Details and Hyperparameters}

\hspace{\parindent} \textbf{ELMo Layer}: We used the official AllenNLP
implementation of the ELMo
model\footnote{\tiny\url{https://allenai.github.io/allennlp-docs/api/allennlp.modules.elmo.html}},
with the official weights pre-trained on the 1 Billion Word Language Model
Benchmark, which contains about 800M tokens of news crawl data from WMT 2011
\cite{Chelba2014}.

\textbf{Dimensionalities}: By default the ELMo layer outputs a 1024-dimensional
vector, which we then feed to a BiLSTM with output size 2048, resulting in a
4096-dimensional vector when concatenating forward and backward directions for
each word of the sequence\footnote{A BiLSTM is composed of two separate LSTMs
that read the input in opposite directions and whose outputs are concatenated at
the hidden dimension. This results in a vector double the dimension of the input
for each time step.}. After max-pooling the BiLSTM output over the sequence
dimension, we obtain a single 4096-dimensional vector corresponding to the tweet
representation. This representation is finally fed to a single-layer
fully-connected network with input size 4096, 512 hidden units, output size 6,
and a ReLU nonlinearity after the hidden layer. The output of the dense layer is
a 6-dimensional logit vector for each input example.

\textbf{Loss Function}: As this corresponds to a multiclass classification
problem (predicting a single class for each example, with more than 2 classes to
choose from), we used the Cross-Entropy Loss as implemented in PyTorch
\cite{paszke2017automatic}.

\textbf{Optimization}: We optimized the model with Adam
\cite{DBLP:journals/corr/KingmaB14}, using default hyperparameters
($\beta_1=0.9$, $\beta_2=0.999$, $\epsilon=10^{-8}$), following a slanted
triangular learning rate schedule \cite{howard2018universal}, also with default
hyperparameters ($cut\_frac=0.1$, $ratio=32$), and a maximum learning rate
$\eta_{max}=0.001$, over $T=23,970$ iterations\footnote{This number is obtained
    by multiplying the number of epochs ($10$), times the total number of
    batches, which for the training dataset corresponds to $2396$ batches of
$64$ elements, and $1$ batch of $39$ elements, hence $2397\times10=23,970$.}.

\textbf{Regularization}: we used a dropout layer \cite{srivastava2014dropout},
with probability of 0.5 after both the ELMo and the hidden fully-connected
layer, and another one with probability of 0.1 after the max-pooling aggregation
layer. We also reshuffled the training examples between epochs, resulting in a
different batch for each iteration. 

\textbf{Model Selection}: To choose the best hyperparameter configuration we
measured the classification accuracy on the validation (trial) set.

\subsection{Ensembles}
 
\begin{figure}[!h]
    \centering
    \includegraphics[width=\columnwidth]{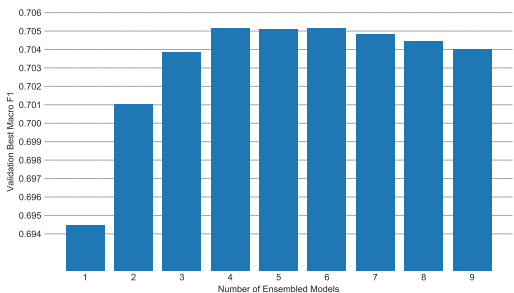}
\caption{Effect of the number of ensembled models on validation performance.}
\label{fig:best_ensembles}
\end{figure}

Once we found the best-performing configuration we trained 10 models using
different random seeds, and tried averaging the output class probabilities of
all their possible $\sum_{k=1}^{9}{\binom{9}{k}}=511$ combinations. As Figure
\ref{fig:best_ensembles} shows, we empirically found that a specific combination
of $6$ models yielded the best results ($70.52\%$), providing evidence for the
fact that using a number of independent classifiers equal to the number of class
labels provides the best results when doing average ensembling
\cite{bonab2016theoretical}.

\section{Experiments and Analyses}

We performed several experiments to gain insights on how the proposed model's
performance interacts with the shared task's data. We performed an ablation
study to see how some of the main hyperparameters affect performance, and an
analysis of tweets containing hashtags and emoji to understand how these two
types of tokens help the model predict the trigger-word's emotion. We also
observed the effects of varying the amount of data used for training the model
to evaluate whether it would be worthwhile to gather more training data.

\subsection{Ablation Study}%
\label{sub:ablation_study}

We performed an ablation study on a single model having obtained 69.23\%
accuracy on the validation set. Results are summarized in Table
\ref{table:ablation}.

\begin{table}[!h]
    \centering
    \footnotesize

    \begin{tabular}{lcc}

        \toprule
        \textbf{Variation} & \textbf{Accuracy (\%)} & $\bf{\Delta}$\textbf{\%} \\
        \midrule[0.08em]
        Submitted          & \textbf{69.23}         & -                        \\
        \midrule[0.05em]
        No emoji           & 68.36                  & - 0.87                   \\
        \midrule[0.05em]
        No ELMo            & 65.52                  & - 3.71                   \\
        \midrule[0.05em]
        Concat Pooling     & 68.47                  & - 0.76                   \\
        \midrule[0.05em]
        LSTM hidden=4096   & 69.10                  & - 0.13                   \\
        LSTM hidden=1024   & 68.93                  & - 0.30                   \\
        LSTM hidden=512    & 68.43                  & - 0.80                   \\
        \midrule[0.05em]
        POS emb dim=100    & 68.99                  & - 0.24                   \\
        POS emb dim=75     & 68.61                  & - 0.62                   \\
        POS emb dim=50     & 69.33                  & + 0.10                   \\
        POS emb dim=25     & 69.21                  & - 0.02                   \\
        \midrule[0.05em]
        SGD optim lr=1     & 64.33                  & - 4.90                   \\
        SGD optim lr=0.1   & 66.11                  & - 3.12                   \\
        SGD optim lr=0.01  & 60.72                  & - 8.51                   \\
        SGD optim lr=0.001 & 30.49                  & - 38.74                  \\
        \bottomrule

    \end{tabular}

    \caption{Ablation study results.}
    \vspace{-0.4cm}
    \caption*{
        \footnotesize Accuracies were obtained from the
        validation dataset. Each model was trained with the same random seed and
        hyperparameters, save for the one listed. ``No emoji'' is the same model
        trained on the training dataset with no emoji, ``No ELMo'' corresponds
        to having switched the ELMo word encoding layer with a simple
        pre-trained GloVe embedding lookup table, and ``Concat Pooling''
        obtained sentence representations by using the pooling method described
        by \citet{howard2018universal}. ``LSTM hidden'' corresponds to the
        hidden dimension of the BiLSTM, ``POS emb dim'' to the dimension of the
        part-of-speech embeddings, and ``SGD optim lr'' to the learning rate
        used while optimizing with the schedule described by \citet{conneau2017supervised}.}

    \label{table:ablation}

\end{table}

We can observe that the architectural choice that had the greatest impact on our
model was the ELMo layer, providing a $3.71\%$ boost in performance as compared
to using GloVe pre-trained word embeddings.

We can further see that emoji also contributed significantly to the model's
performance. In Section \ref{sub:effect_of_emoji_and_hashtags} we give some
pointers to understanding why this is so.  

Additionally, we tried using the concatenation of the max-pooled, average-pooled
and last hidden states of the BiLSTM as the sentence representation, following
\citet{howard2018universal}, but found out that this impacted performance
negatively. We hypothesize this is due to tweets being too short for needing
such a rich representation. Also, the size of the concatenated vector was
$4096\times3=12,288$, which probably could not be properly exploited by the
$512$-dimensional fully-connected layer.

Using a greater BiLSTM hidden size did not help the model, probably because of
the reason mentioned earlier; the fully-connected layer was not big or deep enough
to exploit the additional information. Similarly, using a smaller hidden size
neither helped.

We found that using 50-dimensional part-of-speech embeddings slightly improved results,
which implies that better fine-tuning this hyperparameter, or using a better POS
tagger could yield an even better performance.

Regarding optimization strategies, we also tried using SGD with different
learning rates and a step-wise learning rate schedule as described by
\citet{conneau2018}, but we found that doing this did not improve performance. 

\begin{figure}[!h]
    \centering
    \includegraphics[width=\columnwidth]{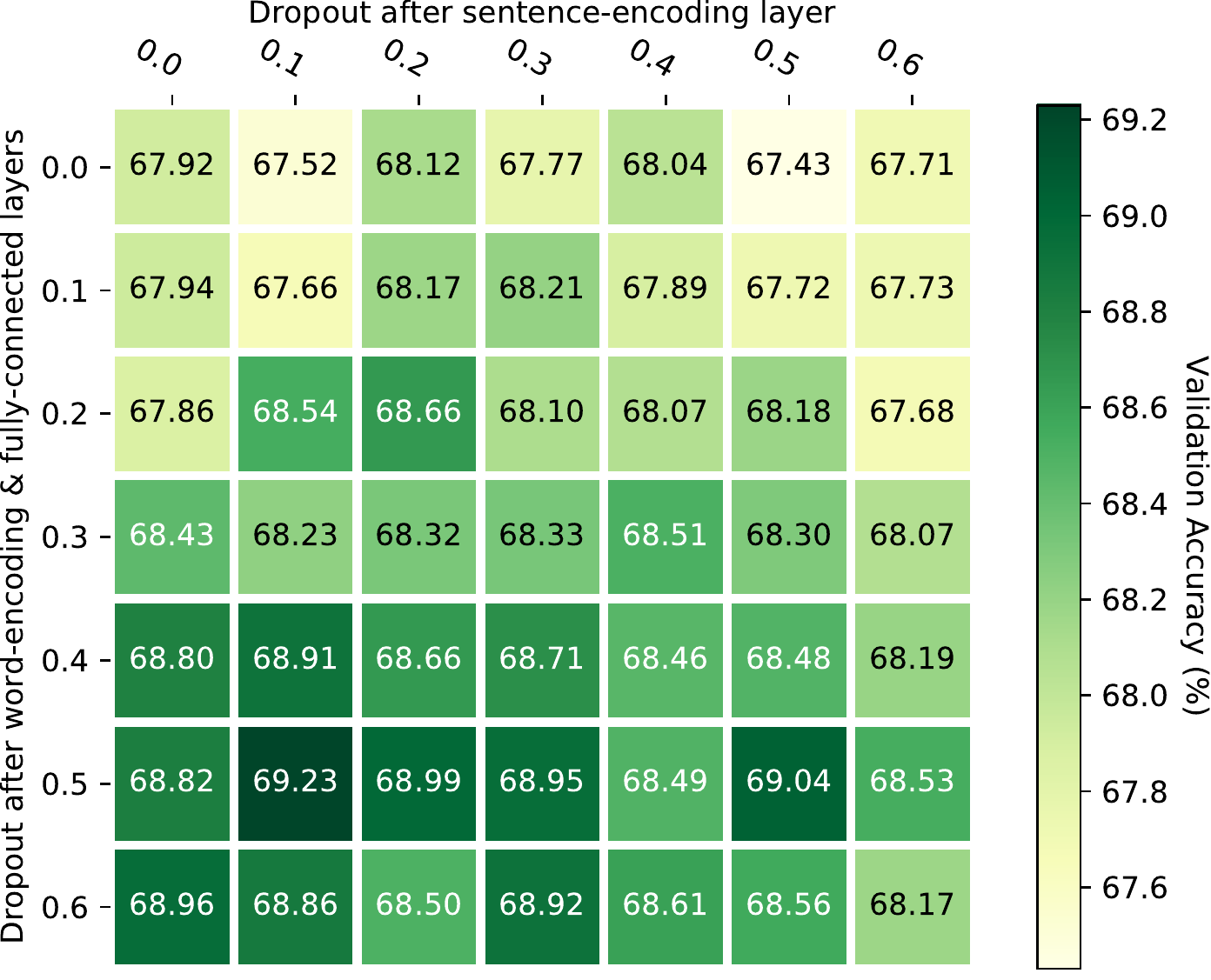}

    \caption{Dropout Ablation.}
    \vspace{-0.4cm}

    \caption*{
        \footnotesize Rows correspond to the dropout applied both after
        the ELMo layer (word encoding layer) and after the fully-connected network's
        hidden layer, while columns correspond to the dropout applied after the
        max-pooling operation (sentence encoding layer.)}

\label{fig:dropouts}
\end{figure}

Finally, Figure \ref{fig:dropouts} shows the effect of using different dropout
probabilities. We can see that having higher dropout after the
word-representation layer and the fully-connected network's hidden layer, while
having a low dropout after the sentence encoding layer yielded better results
overall. 

\subsection{Error Analysis}

\begin{figure}[!h]
    \centering
    \includegraphics[width=\columnwidth]{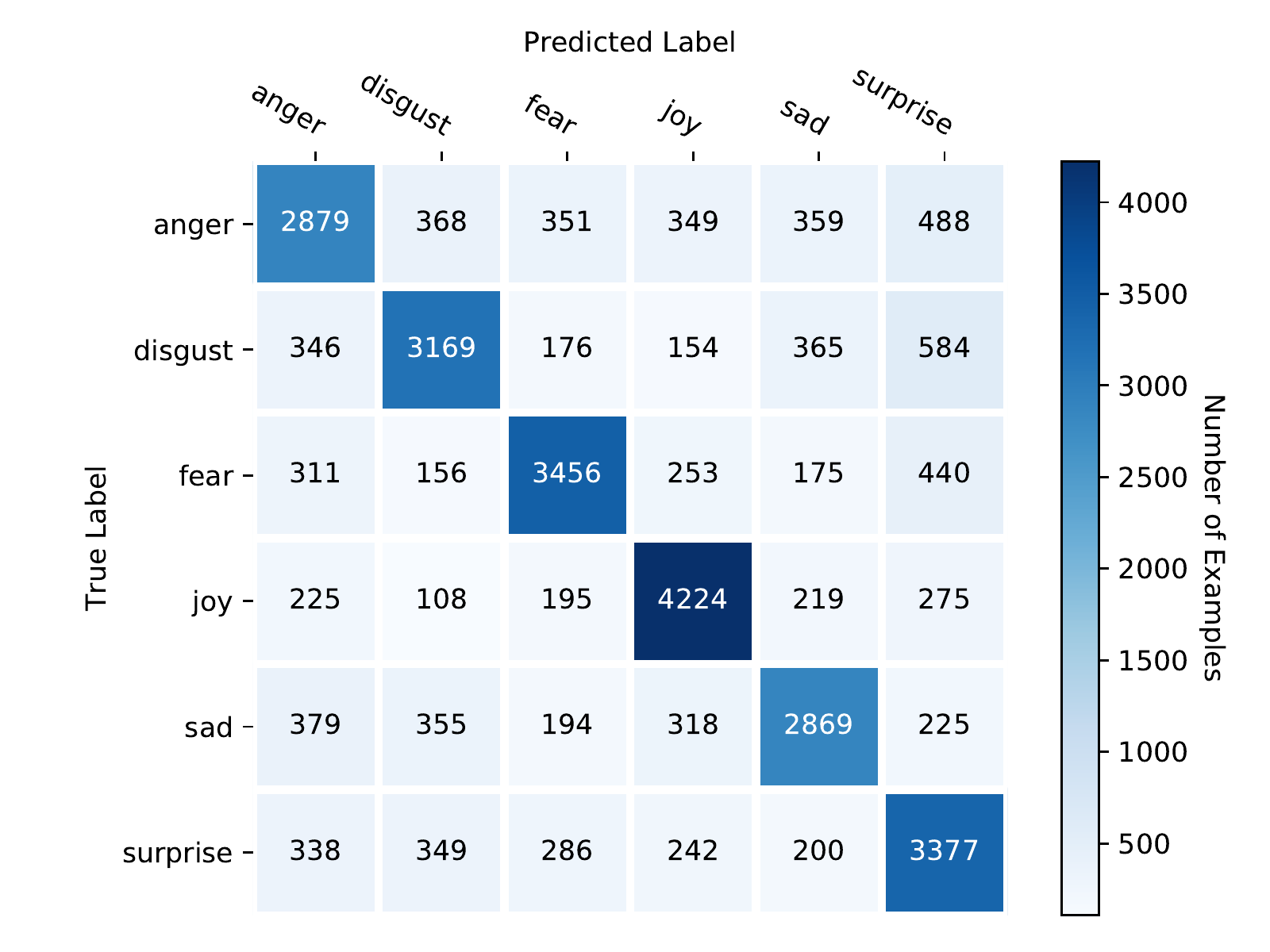}

    \caption{Confusion Matrix (Test Set).}
    % \vspace{-0.4cm}

    % \caption*{
    %     \footnotesize Rows correspond to the dropout applied both after
    %     the ELMo layer (word encoding layer) and after the fully-connected network's
    %     hidden layer, while columns correspond to the dropout applied after the
    %     max-pooling operation (sentence encoding layer.)}

    \label{fig:confusion_matrix}
\end{figure}

Figure~\ref{fig:confusion_matrix} shows the confusion matrix of a single model
evaluated on the test set, and Table~\ref{table:classification_report} the
corresponding classification report. In general, we confirm what
\newcite{klinger2018iest} report: \texttt{anger} was the most difficult class to
predict, followed by \texttt{surprise}, whereas \texttt{joy}, \texttt{fear}, and
\texttt{disgust} are the better performing ones.

\begin{figure}[!h]
    \centering
    \includegraphics[width=\columnwidth]{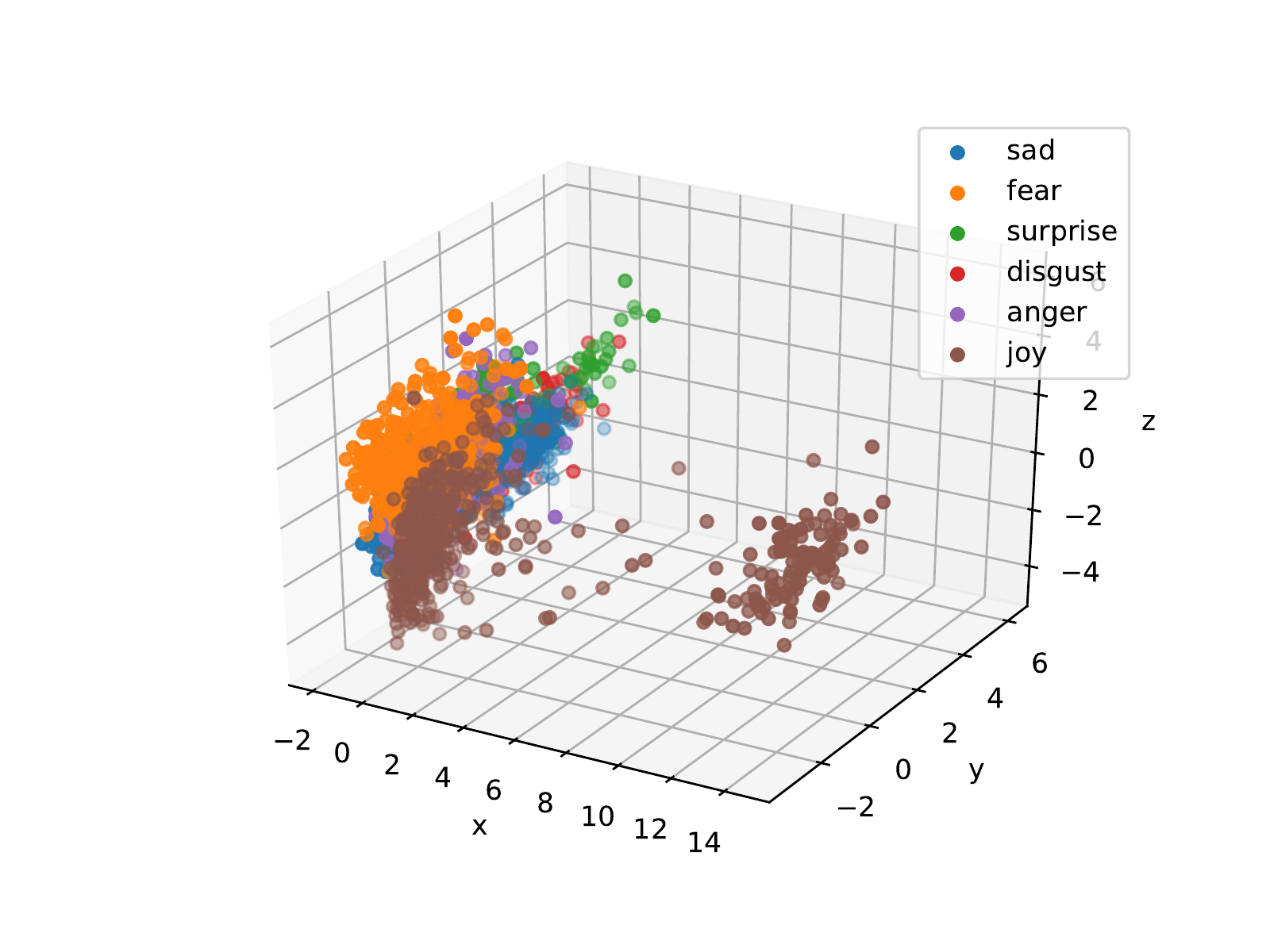}

    \caption{3d Projection of the Test Sentence Representations.}
    \label{fig:pca}
\end{figure}

To observe whether any particular pattern arose from the sentence
representations encoded by our model, we projected them into 3d space through
Principal Component Analysis (PCA), and were surprised to find that 2 clearly
defined clusters emerged (see Figure \ref{fig:pca}), one containing the majority
of datapoints, and another containing \texttt{joy} tweets exclusively. Upon
further exploration we also found that the smaller cluster was composed only by
tweets containing the pattern \texttt{un~\_\_TRIGGERWORD\_\_}, and further,
that all of them were correctly classified.

It is also worth mentioning that there are 5827 tweets in the training set with
this pattern. Of these, 5822 (99.9\%) correspond to the label \texttt{joy}. We
observe a similar trend on the test set; 1115 of the 1116 tweets having the
\texttt{un~\_\_TRIGGERWORD\_\_} pattern correspond to \texttt{joy} tweets. We
hypothesize this is the reason why the model learned this pattern as a strong
discriminating feature.

Finally, the only tweet in the test set that contained this pattern and did not
belong to the \texttt{joy} class, originally had \textit{unsurprised} as its
triggerword\footnote{We manually searched for the original tweet.}, and
unsurprisingly, was misclassified.

\begin{table}[htpb]
    \centering
    \footnotesize
    
    \begin{tabular}{cccc}

        \toprule
                          & \textbf{Precision} & \textbf{Recall} & \textbf{F1-score} \\
        \midrule
        \texttt{anger}    & 0.643     & 0.601  & 0.621    \\
        \texttt{disgust}  & 0.703     & 0.661  & 0.682    \\
        \texttt{fear}     & 0.742     & 0.721  & 0.732    \\
        \texttt{joy}      & 0.762     & 0.805  & 0.783    \\
        \texttt{sad}      & 0.685     & 0.661  & 0.673    \\
        \texttt{surprise} & 0.627     & 0.705  & 0.663    \\
        \midrule
        Average           & 0.695     & 0.695  & 0.694    \\
        \bottomrule
    \end{tabular}

    \caption{Classification Report (Test Set).}

\label{table:classification_report}
\end{table}

\subsection{Effect of the Amount of Training Data}

\begin{figure}[!h]
    \centering
    \includegraphics[width=\columnwidth]{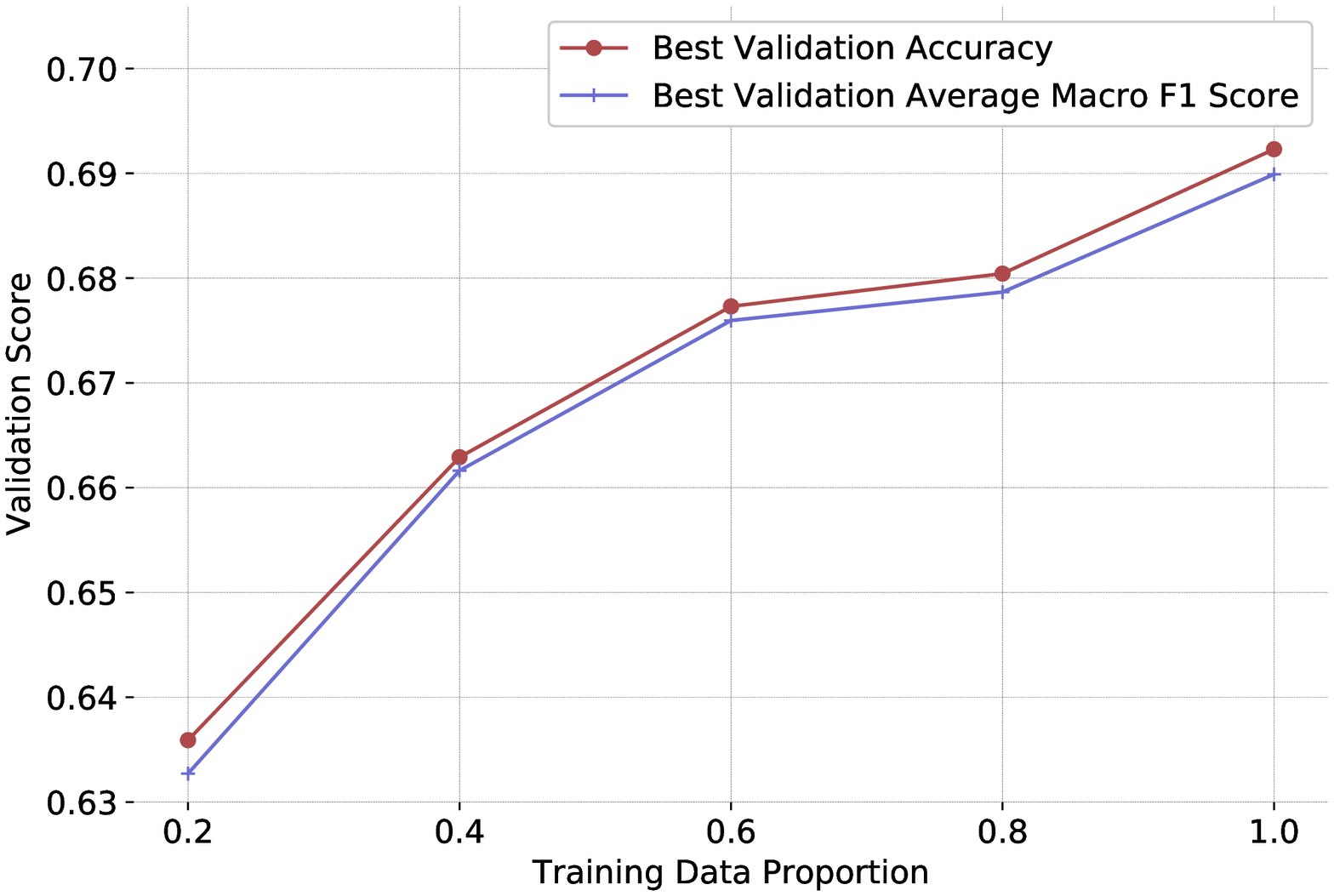}
    \caption{Effect of the amount of training data on classification performance.}
    \label{fig:data_amt_vs_acc}
\end{figure}

As Figure \ref{fig:data_amt_vs_acc} shows, increasing the amount of data with
which our model was trained consistently increased validation accuracy and
validation macro F1 score. The trend suggests that the proposed model is
expressive enough to learn from more data, and is not overfitting the training
set. 

\subsection{Effect of Emoji and Hashtags}%
\label{sub:effect_of_emoji_and_hashtags}

\begin{table}[!h]
    \centering
    \footnotesize

    \begin{tabular}{lcc}

        \toprule
        \textbf{} & \textbf{Present} & \textbf{Not Present} \\
        \midrule
        Emoji     & 4805 (76.6\%)    & 23952 (68.0\%)       \\
        Hashtags  & 2122 (70.5\%)    & 26635 (69.4\%)       \\
        \bottomrule

    \end{tabular}

    \caption{Number of tweets on the test set with and without emoji and
    hashtags. The number between parentheses is the proportion of tweets
    classified correctly.}

    \label{table:emoji_and_hashtags}

\end{table}

Table \ref{table:emoji_and_hashtags} shows the overall effect of hashtags and
emoji on classification performance. Tweets containing emoji seem to be easier
for the model to classify than those without. Hashtags also have a positive
effect on classification performance, however it is less significant. This
implies that emoji, and hashtags in a smaller degree, provide tweets with a
context richer in sentiment information, allowing the model to better guess the
emotion of the \textit{trigger-word}.

\begin{table}[!h]
    \centering
    \scriptsize
        \begin{tabular}{lrrcrcr}

            \toprule
           \multirow{2}{*}{\textbf{Emoji alias}} & \multirow{2}{*}{\textbf{N}\hspace{0.1cm}} & \multicolumn{2}{c}{\textbf{emoji}} & \multicolumn{2}{c}{\textbf{no-emoji}} &
            \multirow{2}{*}{$\bf{\Delta}$\textbf{\%}} \\
            \cmidrule(lr){3-4}
            \cmidrule(lr){5-6}
                                &     & \#   & \% & \# & \% & \\
            \midrule[0.08em]
            \texttt{mask}       & 163   \hspace{0.1cm} & 154 & 94.48 \hspace{0.1cm} & 134 & 82.21 & - 12.27 \\
          \texttt{two\_hearts}  & 87    \hspace{0.1cm} & 81  & 93.10 \hspace{0.1cm} & 77  & 88.51 & - 4.59  \\
          \texttt{heart\_eyes}  & 122   \hspace{0.1cm} & 109 & 89.34 \hspace{0.1cm} & 103 & 84.43 & - 4.91  \\
          \texttt{heart}        & 267   \hspace{0.1cm} & 237 & 88.76 \hspace{0.1cm} & 235 & 88.01 & - 0.75  \\
          \midrule[0.001em]
          \texttt{rage}         & 92    \hspace{0.1cm} & 78  & 84.78 \hspace{0.1cm} & 66  & 71.74 & - 13.04 \\
          \texttt{cry}          & 116   \hspace{0.1cm} & 97  & 83.62 \hspace{0.1cm} & 83  & 71.55 & - 12.07 \\
          \texttt{sob}          & 490   \hspace{0.1cm} & 363 & 74.08 \hspace{0.1cm} & 345 & 70.41 & - 3.67  \\
          \texttt{unamused}     & 167   \hspace{0.1cm} & 121 & 72.46 \hspace{0.1cm} & 116 & 69.46 & - 3.00  \\
          \midrule[0.001em]
          \texttt{weary}        & 204   \hspace{0.1cm} & 140 & 68.63 \hspace{0.1cm} & 139 & 68.14 & - 0.49  \\
          \texttt{joy}          & 978   \hspace{0.1cm} & 649 & 66.36 \hspace{0.1cm} & 629 & 64.31 & - 2.05  \\
          \texttt{sweat\_smile} & 111   \hspace{0.1cm} & 73  & 65.77 \hspace{0.1cm} & 75  & 67.57 & 1.80 \\
          \texttt{confused}     & 77    \hspace{0.1cm} & 46  & 59.74 \hspace{0.1cm} & 48  & 62.34 & 2.60 \\
          \bottomrule

        \end{tabular}

    \caption{Fine grained performance on tweets containing emoji, and the effect
of removing them.} 
    \vspace{-0.4cm}
    \caption*{
        \footnotesize \textbf{N} is the total number of tweets containing the listed emoji,
        \textbf{\#} and \textbf{\%} the number and percentage of correctly-classified
        tweets respectively, and $\bf{\Delta}$\textbf{\%} the variation of test accuracy
        when removing the emoji from the tweets.}

    \label{table:emoji_fine_grained}

\end{table}

Table \ref{table:emoji_fine_grained} shows the effect specific emoji have on
classification performance. It is clear some emoji strongly contribute to improving prediction quality. The
most interesting ones are \texttt{mask}, \texttt{rage}, and \texttt{cry}, which
significantly increase accuracy. Further, contrary to intuition, the
\texttt{sob} emoji contributes less than \texttt{cry}, despite representing a
stronger emotion. This is probably due to \texttt{sob} being used for depicting
a wider spectrum of emotions.

Finally, not all emoji are beneficial for this task. When removing
\texttt{sweat\_smile} and \texttt{confused} accuracy increased, probably because
they represent emotions other than the ones being predicted.

%From observing Table \ref{table:emoji_label} it seems likely that tweets
%containing \texttt{mask} would be highly predictive of the \texttt{disgust}
%class, since more than $92\%$ of the tweets containing such emoji belong to
%that class, however it less clear why \texttt{rage}, and \texttt{cry}
%contribute similarly, despite being more evenly distributed among classes.

%This means that they are highly representative of the the class the tweet
%belongs to, and would be good choices for discriminating new tweets in case we
%wanted to expand the dataset.

%For example, by observing Table \ref{table:emoji_label} we can infer that
%\texttt{mask}, \texttt{rage}, and \texttt{cry}   \texttt{disgust},
%\texttt{rage}, and \texttt{sad} classes respectively.

\section{Conclusions and Future Work}

We described the model that got second place in the WASSA 2018 Implicit Emotion
Shared Task. Despite its simplicity, and low amount of dependencies on libraries
and external features, it performed almost as well as the system that obtained
the first place.

Our ablation study revealed that our hyperparameters were indeed quite
well-tuned for the task, which agrees with the good results obtained in the
official submission. However, the ablation study also showed that increased
performance can be obtained by incorporating POS embeddings as additional
inputs. Further experiments are required to accurately measure the impact that
this additional input may have on the results. We also think the performance can
be boosted by making the architecture more complex, concretely, by using a
BiLSTM with multiple layers and skip connections in a way akin to
\cite{peters2018deep}, or by making the fully-connected network bigger and
deeper.

We also showed that, what was probably an annotation artifact, the
\texttt{un~\_\_TRIGGERWORD\_\_} pattern, resulted in increased performance for
the \texttt{joy} label. This pattern was probably originated by a heuristic
naïvely replacing the ocurrence of \textit{happy} by the trigger-word indicator.
We think the dataset could be improved by replacing the word \textit{unhappy},
in the original examples, by \texttt{\_\_TRIGGERWORD\_\_} instead of
\texttt{un~\_\_TRIGGERWORD\_\_}, and labeling it as \texttt{sad}, or
\texttt{angry}, instead of \texttt{joy}.

% To prevent future models from mistakenly learning this
% feature as a strong indicator for such label, it would be wise to add more
% examples from other classes cotaining the same pattern, or removing them
% entirely and rebalancing the dataset.

Finally, our studies regarding the importance of hashtags and emoji in the
classification showed that both of them seem to contribute significantly to the
performance, although in different measures. 

\section{Acknowledgements}%
\label{sec:acknowledgements}

We thank the anonymous reviewers for their reviews and suggestions. The first
author is partially supported by the Japanese Government MEXT Scholarship.

\bibliography{bibliography}
\bibliographystyle{acl_natbib_nourl}

\end{document}